\documentclass[1p, review]{elsarticle} %单栏
\usepackage{amssymb}
\usepackage[numbers]{natbib}
\usepackage{amssymb}
\usepackage{graphicx}
\usepackage{bbm}
\usepackage{booktabs}
\usepackage{amsmath}
\usepackage{algorithmic}
\usepackage{array}
\usepackage{lineno}

\journal{}

\begin{document}
%\linenumbers

\begin{frontmatter}



\title{Discovery-and-Selection$:$ Towards Optimal Multiple Instance Learning for Weakly Supervised Object Detection}


\author[address1]{Shiwei Zhang}
\ead{zhangshiwei96@stu.xjtu.edu.cn}

\author[address1]{Lin Yang}
\ead{3120305430@stu.xjtu.edu.cn}

\author[address1]{Wei Ke\corref{cor1}}
\ead{wei.ke@mail.xjtu.edu.cn}
\cortext[cor1]{Corresponding author}%corresponding author
\address[address1]{School of Software Engineering, Xi’an Jiaotong University, Xi'an, Shaanxi, China}

\begin{abstract}
Weakly supervised object detection (WSOD) is a challenging task that requires simultaneously learn object classifiers and estimate object locations under the supervision of image category labels. A major line of WSOD methods roots in multiple instance learning which regards images as bags of instances and selects positive instances from each bag to learn the detector. However, a grand challenge emerges when the detector inclines to converge to discriminative parts of objects rather than the whole objects. In this paper, under the hypothesis that optimal solutions are included in local minima, we propose a discovery-and-selection approach fused with multiple instance learning (DS-MIL), which finds rich local minima and select optimal solution from multiple local minima. To implement DS-MIL, an attention module is proposed so that more context information can be captured by feature maps and more valuable proposals can be collected during training. With proposal candidates, a selection module is proposed to select informative instances for object detector. Experimental results on commonly used benchmarks show that our proposed DS-MIL approach can consistently improve the baselines, reporting state-of-the-art performance.
\end{abstract}

\begin{keyword}
Weakly Supervised Object Detection \sep Training Strategy  \sep Self-Attention 
\end{keyword}

\end{frontmatter}

\section{Introduction}
\label{sec:introduction}
\begin{figure*}[!ht]
\centering
\includegraphics[scale=0.45]{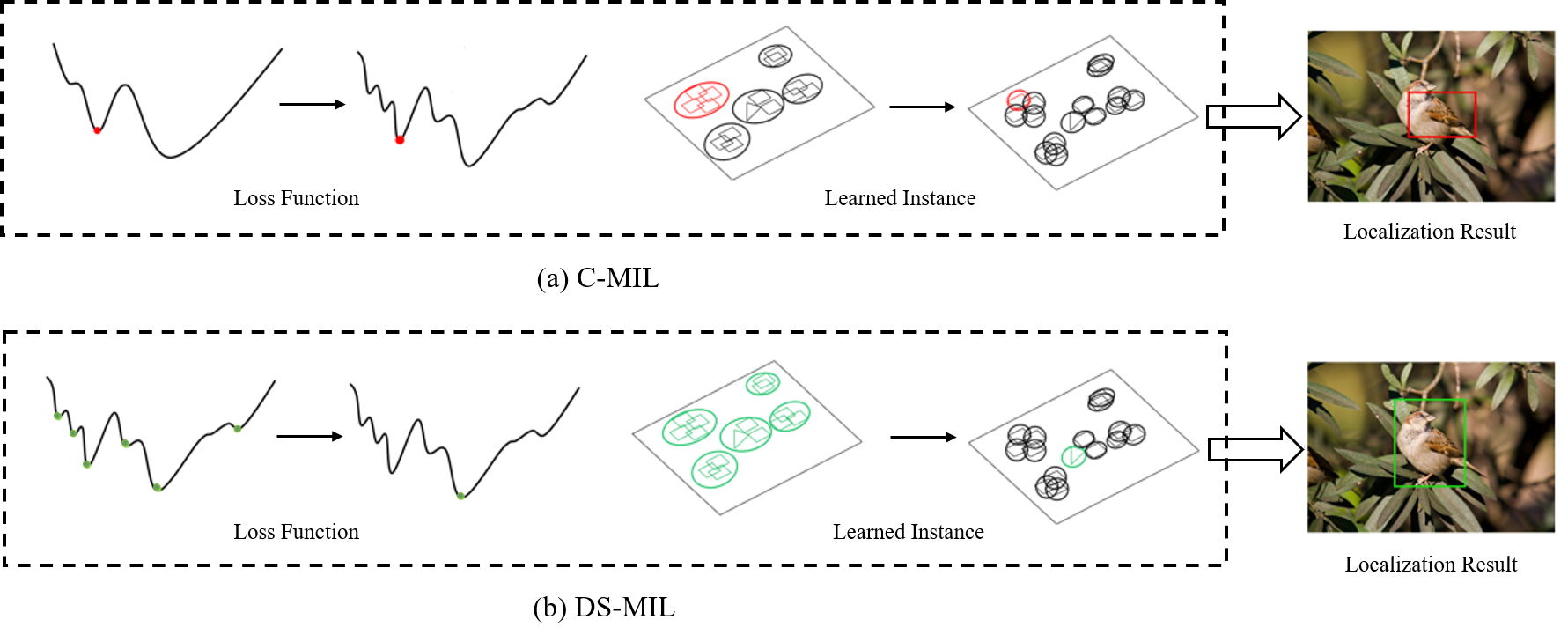}
\caption{Comparison of CMIL based approaches and our DS-MIL. (a) shows that C-MIL introduced the continuation method into WSOD, but still can not solve the non-convexity problem completely and localize only part of object (Red box).   (b) shows our motivation which introduce a Discovering Module to find more local-minima and a Selection Module to choose in the found instances. This alleviates the non-convexity problem and localizes full object extent (Green box).}
\label{fig:1}
\end{figure*}
Weakly supervised object detection (WSOD) has been attracted increasing attention, due to its effortless annotation that only needs indicator vectors to demonstrate the existence of each class~\cite{Bilen_2016_CVPR, diba2017weakly, tang2017multiple, tang2018pcl, wan2018min, wan2019c, huang2020comprehensive}. Compared with fully supervised object detection which requires labor-intensive bounding-box annotations, WSOD significantly reduces the workload of data annotation. With WSOD, people can leverage rich images with tags on the internet to learn object-level models, and thereby convert human-supervised object detection to Weakly supervised object modeling.

Multiple Instance Learning (MIL)~\cite{DBLP:conf/nips/MaronL97} has been the cornerstone of many WSOD methods, either with hand-crafted features~\cite{DBLP:conf/cvpr/HoffmanPDS15,DBLP:journals/pami/CinbisVS17} or deep learning pipelines~\cite{Bilen_2016_CVPR,tang2017multiple,tang2018pcl,wan2019c,ren2020instance}. 
With MIL, images are decomposed to bags of proposals (instances). Each image from the classes of interest has at least one positive instance and images from negative classes has no positive instance.
WSOD is considered as an instance classification problem, where object detectors are constructed by alternating training the classifier and selecting positive candidate proposal.

MIL-based WSOD networks usually focus on classifier learning and feature learning, which roughly choose the high-scored candidate as positive samples for the object localization. Consequently, the detectors rely on classification score outputted by the MIL classifier, resulting in noisy proposals of poor localization. 
The noisy proposals are typically discriminative object parts instead of whole object extent.

%\
%
To alleviate the impact of noisy proposals, one solution is re-training an object detector with pseudo ground-truths (top-scoring proposals) generated by weakly-supervised object detectors~\cite{li2016weakly,tang2017multiple,tang2018pcl,wan2019c}. 
However, because the number of  the noisy proposals are usually greater than the optimal solution, the noisy proposal introduced in the training phase could seriously deteriorate the trained detectors.The other solution is to explore sophisticated optimization strategies. The C-MIL method~\cite{wan2019c} recognized this problem by decomposing the complicated optimization problem to multiple sub-optimization problems which are easy to be solved. Nevertheless, as shown in Fig.~\ref{fig:1}(a), C-MIL remains getting stuck to the local minimum when the continuation parameters are not properly defined.
%
%But this method is still have weakness. As shown in Fig. \ref{fig:1}, despite introducing continuation method, it is still stuck into local minimum because the continuation parameters are hard to choose. 

In this paper, we introduce a discovery-and-selection training strategy in Fig.~\ref{fig:1}(b) to multiple instance learning network and thereby create DS-MIL.
DS-MIL is implemented by introducing an instance discovery module and an instance selection module to the multiple instance learning network.
It aims to discover multiple local minima and then select the best sample in these multiple local minima, alleviating the local minimum issue in WSOD in an simple-yet-effective fashion.
%we propose a discovery-and-selection training strategy for WSOD, with the aim to address the non-convexity problem and avoid localizing object parts.
%

{
For the discovery term, inspired by non-local network~\cite{wang2018non}, a self-attention module is designed so that the feature maps of CNN capture context information of the object proposals generated by Selective Search. 
In this manner, we can find rich local minima, which increases the probability to obtain optimal solution during multiple instance learning. 
% For the selection term, we take an Expectation-Maximization algorithm to re-rank the confidence of the object proposals, in which we explicitly model instance assignment as a hidden variable and derive the pseudo-label generation scheme to conduct the E and M steps respectively. 
For the selection term, %we re-rank the confidence of the object proposals, in which we explicitly model instance assignment as a hidden variable and combine the proposals' importance and probability of the class of the proposals to classify the image. 
%The algorithm assign a high score to the proposals which lays a decisive role to determine an proposal bag whether belongs to positive.
because the label of the image is triggered by the key instances' label, we propose selection module to get proposals importance and proposals' category. Then we combine both of them to predict the category of the image which can be supervised by the image-level label. In the end, after adequate training, the proposals importance we got can help us to choose the most important proposal from rich local minima outputted by dicovery module.

}

The contributions of this paper are summarized as follows:
{
\begin{enumerate}
    \item We propose the discovery-and-selection training strategy for WSOD, solving the local minimum issue of multiple instance learning under the hypothesis that optimal solution is included in local minima.
    
    \item We design a proposal discovery module which leverages localization information from multiple locations and finds more reliable proposals. 
    We propose a novel proposal selection module to optimize instance proposals. 
    
    \item Experimental results on commonly used benchmarks show our proposed DS-MIL approach can consistently improve the baselines, achieving the state-of-the-art performances.
\end{enumerate}
}

The rest of this paper is organized as follows: In Section \uppercase\expandafter{\romannumeral2}, we review related research. In Section \uppercase\expandafter{\romannumeral3} we describe the proposed approach in details. Experimental results are shown and discussed in Section \uppercase\expandafter{\romannumeral4}, and we made a conclusion of our work in Section \uppercase\expandafter{\romannumeral5}.

\section{Related Works}
\label{sec:relatedwork}

WSOD is an attractive computer vision task in which a detector is trained only with image-level annotations. WSOD is usually solved with MIL based approach, especially significantly boosted with convolutional neural networks.  

\subsection{Multiple Instance Learning for WSOD.}

MIL is effective to solve weakly supervised problem with coarse labels~\cite{DBLP:conf/nips/MaronL97}. Positive and negative bags are used to train a instance-level classifier in MIL. 
A positive bag is a set of instances at least one of which is positive while a negative bag is a set of negative instances. 
The WSOD is natural to treat as a MIL problem. Supposing image is a bag with candidate instances which are generated by object proposal method~\cite{uijlings2013selective}.
The multi-fold MIL is proposed to solve large-scale training dataset by diving it to several parts~\cite{DBLP:conf/cvpr/HoffmanPDS15}. In~\cite{DBLP:journals/pami/CinbisVS17}, full annotation of extra data is used to train a instance detector, improving the performance of MIL by transferring representation.
However, the performance gap between weakly supervised and fully supervised task is insurmountable with traditional MIL approaches.

\subsection{Deep Learning for WSOD}

Recently, WSOD largely outperforms the previous State-of-the-arts by combining deep neural networks and MIL. 
The Weakly Supervised Deep Detection (WSDDN)~\cite{Bilen_2016_CVPR} is firstly introduced to deal with WSOD, which is composed of a proposal classifier branch and a proposal selector branch inspired by MIL. WSDDN selects the positive samples by aggregating the score of the two branches and its effectiveness attracts lots of works to follow its framework. The WSDDN brings the WSOD into a new era.

\textit{Feature Learning based WSOD.}
\cite{singh2016track} transfered tracked boxes from weakly-labeled videos to weakly-labeled images as pseudo ground-truth to train the detector directly on images.
\cite{ge2018multi} proposed to fuse and filter object instances from different techniques and perform pixel labeling with uncertainty and they used the resulting pixel-wise labels to generate bounding boxes for object detection and attention maps for multi-label classification.
Others are attempt to learn feature representation to gain better performance. \cite{diba2017weakly} proposed an end-to-end cascaded convolutional network to perform weakly supervised object detection and segmentation in cascaded manner.
\cite{kantorov2016contextlocnet} proposed to learn a context-aware CNN with contrast-based contextual modeling.
\cite{choe2019attention} uses mask to hide the most discriminative part of a image to enforce the feature extractor to capture the integral extent of object. 
\cite{shen2019cyclic} leverage the complementary effects of WSOD and Weakly Supervised Semantic Segmentation to build detection-segmentation cyclic collaborative frameworks. 
Comprehensive Attention Self-Distillation (CASD) is proposed to balance feature learning among all object instances\cite{huang2020comprehensive}. 
\cite{wan2018min} inspired by a classical thermodynamic principle, proposed a min-entropy latent model (MELM) and recurrent learning algorithm for weakly supervised object detection.

\textit{Proposal Refinement based WSOD.}
Several approaches focus on the refinement of proposal localization. \cite{li2016weakly} introduces domain adaptation into WSOD to fine-tune the network to collect class specific object proposals. In~\cite{tang2017multiple}, Online Instance Classifier Refinement (OICR) alleviates the part domination problem by knowledge distillation. \cite{tang2018pcl} is based on OICR, coming up with using proposal clustering to improve proposal generation and using proposal clusters as supervision. In order to generate more precise proposals for detection. \cite{tang2018weakly} designed a weakly supervised region proposal network, \cite{wei2018ts2c} proposed a tight box mining method that leverages surrounding segmentation context derived from weakly supervised segmentation to suppress low quality distracting candidates and boost the high-quality ones. \cite{DBLP:journals/pami/CinbisVS17} proposed a multi-fold MIL detector by re-labeling proposals and retraining the object classifier iteratively to prevent the detector from being locked into inaccurate object locations. \cite{zhang2018w2f} proposed a pseudo label excavation algorithm and a pseudo label adaptation algorithm to refine the pseudo labels obtained by \cite{tang2017multiple}. \cite{GAO2022108233} proposed a collaborative module to fuse multiple MIL learners outputs to generate precise object localization. \cite{ZHANG201868} proposed PGE and PGA algorithm to mine and refine pseudo ground truths.
~\cite{zeng2019wsod2,gao2018c,ren2020instance} integrate bounding box regressor into weakly-supervised detector.  
\cite{gao2019c} leverage weakly supervised semantic segmentation to remove unprecise proposals.

\textit{Optimization Strategy for WSOD.}
\cite{liu2019utilizing} observes that the result of MIL based detector is unstable when different initialization is used and it utilizes the instability to improve the performance of the detector by fusing the results of differently initialized detectors.
C-MIL~\cite{wan2019c} is proposed in order to alleviate the non-convexity problem by introducing continuation learning to WSOD to simplify the original MIL loss function. 
\cite{jie2017deep} proposed a self-taught learning approach to progressively harvest high-quality positive instances. 
\cite{li2019weakly} introduces a generative adversarial segmentation module interacts with the conventional detection module to avoid being trapped in local-minima.

\subsection{Weakly Supervised Video Action Localization}
Similar with the setting of WSOD, Weakly Supervised Video Action Localization aims to localize and classify the activities in an untrimmed video with only action label to identify the video has what kind of actions. \cite{wang2017untrimmednets,nguyen2018weakly} uses attention in their methods to compute the importance of each clip. In order to localize complete activities, some adversarial methods\cite{singh2017hide,zeng2019breaking} mask the most conspicuous part of videos.\cite{Liu_2019_CVPR} uses a prior that motionless video clips are unlikely to be actions to separate action clips from complex background. \cite{duchenne2009automatic,gan2016webly,ding2018weakly} try to use other weak labels such as scripts,images from web or action lists to train their model. 
\cite{luo2020weakly} adopts Expectation-Maximization to make the video proposal selection more accuracy. Inspired by \cite{luo2020weakly}, we take the same selection strategy for object proposal selection, which also shows effectiveness for WSOD.  

\subsection{Attention in Object Detection.}
\begin{figure*}[!ht]
\centering
\includegraphics[scale=0.65]{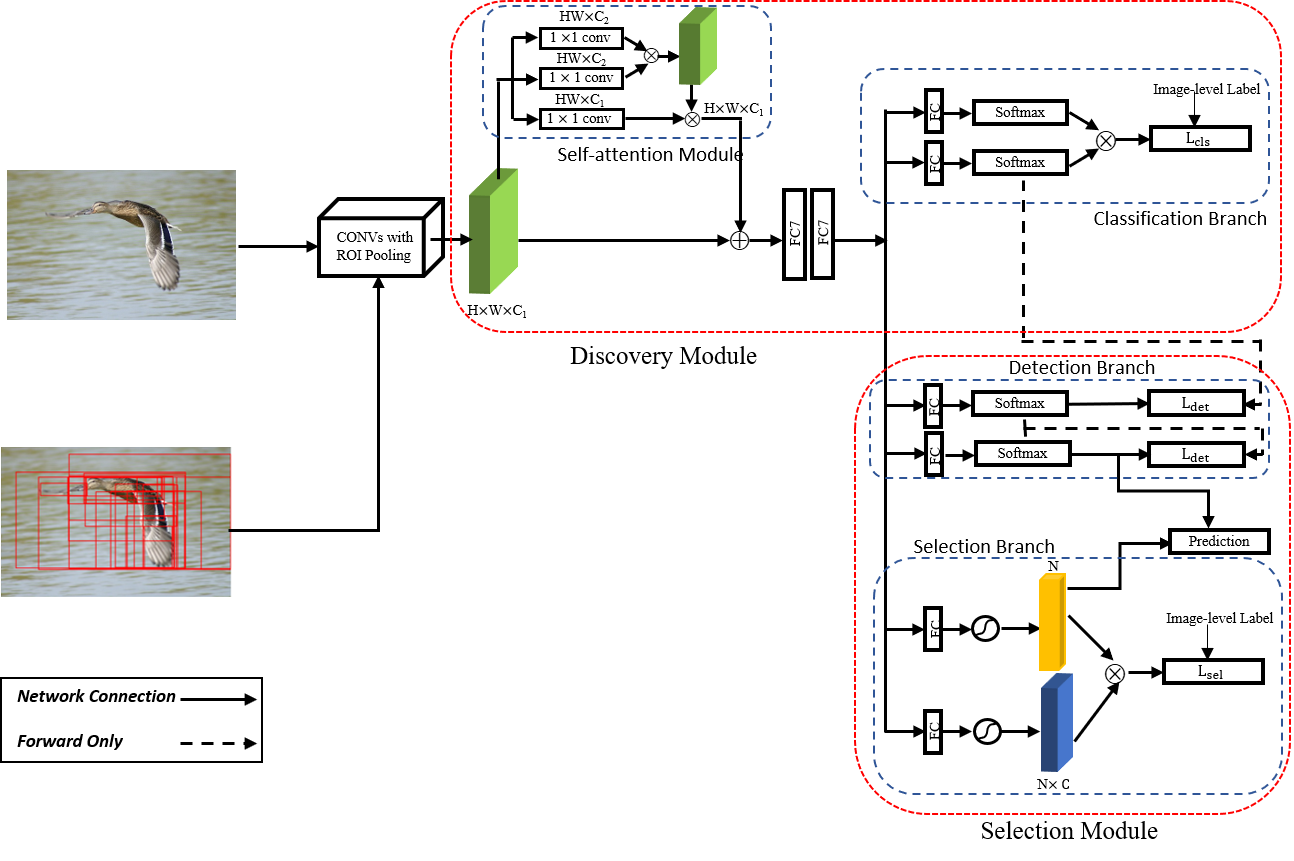}
\caption{The Architecture of DS-MIL Network. (1) Discovering Module is composed by Self-attention branch and classification branch. This module can generate comprehensive features and find more valuable proposals. (2) Selection Module is composed by detection branch and selection branch. This module is proposed to make a selection in all of found proposals.} %(3) Classification branch: Generate pseudo labels for the first detection branch by MIL network. (4) Detection branch: Feed the extracted features and generated pseudo labels to the next regression branch for proposal classification and regression.} %(4) Propoosal Attention generate more accurate score of proposals to re-weight the scores generated by regression branch. In E step, we depend on current parameters to pick key proposals for each positive bag. In M step we use key proposals to update the parameters.}
\label{fig:2}
\end{figure*}
Inspired by the process that humans selectively use an important part of the data to make a decision, attention mechanism was first proposed to solve natural language processing problems and then introduced to computer vision areas\cite{itti2001computational,larochelle2010learning}.  
For object detection, attention mechanism could be classified into two categories:  features re-weighting ~\cite{woo2018cbam,fu2019dual,hu2018squeeze} and loss regularizing~\cite{huang2020improving,ke2020multiple}.
Attention is called self-attention when query is set as itself. Several previous works, $i.e.$, non-local attention~\cite{wang2018non} and relation attention~\cite{hu2018relation}, indicate that self-attention is effective to learn the meaningful representation for conducting the given task. 
We attempt to optimizes the location and classification in WSOD by using both self-attention to explore channel-wise feature re-weighting and normal attention for proposal-wise loss regularization.

It's worth exploring how to effectively take the complementary of the feature learning and proposal selection. By incorporating the attention mechanism, we propose discovering-and-selection strategy, which towards optimal multiple instance learning for weakly supervised object detection.

\section{Method}
\label{sec:method}
We first revisit the most popular method in WSOD, the MIL-based method, which consists of a classification branch and a detection branch. However, previous MIL-based WSOD methods also meet some serious challenges, \textit{i.e.}, the classification branch usually stuck into local minima so that can not generate accurate pseudo labels, and the detection branch also perform poorly because the noisy proposals. Therefore, we propose the DS-MIL to solve this problem as far as possible. As shown in Fig.~\ref{fig:2}, DS-MIL consists of two parts: Discovery Module and Selection Module. The discovery module plug a self-attention module before the classification branch in order to discover rich local minima and generate accurate labels. The selection module add a selection branch to detection branch so that we can re-rank proposals and select the best sample in local minima provided by discovery module.  

\subsection{Revisiting MIL-based WSOD}

MIL-based WSOD model usually follows a two-phase learning procedure, $i.e.$, Classification branch and Detection branch for refinement and regression. It denotes $I=\{I^1, I^2,...,I^T\}$ as an dataset with $T$ images and $Y=\{Y^1, Y^2,...,Y^T\}$ indicating object presence or not. 
Different from fully supervised object annotation with both location and category, the $Y^t=[y^t_1, y^t_2, ..., y^t_C ] \in [0, 1]^{C}$ is a binary vector where $y^t_c=1$ indicates the presence of at least one object of the $c$-th category, where $C$ indicates the total number of object categories in the dataset. 

Suppose $R^t$ is the candidate proposals for the $t$-th image. Each image is pre-computed by Selective Search\cite{uijlings2013selective} to generate $N$ object proposals $R^t=\{R_t^1,R_t^2,...,R_t^N\}$ for initialization. The selected proposals $r$ is a latent variable and $R^t$ can be regarded as the solution space. 
Denoting $\delta$ as the network parameters, the MIL model with proposal selection $r^\ast$ and features $\delta^\ast$ to be learned, can be defined as 
\begin{equation}
\begin{aligned}
     \{r^\ast,\delta^\ast\}&=\mathop{\arg\min}_{r,\delta} \mathcal{L}_{(I,Y)}(r,\delta)\\
     &= \mathop{\arg\min}_{r,\delta}(\mathcal{L}_{cls}+\mathcal{L}_{det})
\end{aligned}
\quad,
\end{equation}
where the image index $t$ are omitted for short and $\mathcal{L}_{cls}$ and $\mathcal{L}_{det}$  are the loss functions of instance classification and proposal detection respectively. 

\textbf{Classification Branch.} Initially, for instance classification term, the loss function is defined as
\begin{equation}
\begin{aligned}
    \mathcal{L}_{cls} = - \sum_{c=1}^C\{y_c \log p_c(r; \delta)+(1-y_c)\log(1-p_c(r; \delta))\},\\
\end{aligned}
\end{equation}
where $p_c(r; \delta)$ is the joint probability of class $c$ and latent variable $r$, given learned network parameter $\delta$. It is calculated by a soft-max operation with the prediction score $s(r; \delta)$, as
\begin{equation}
\begin{aligned}
    p_c(r; \delta)=\dfrac{exp(s(r; \delta))}{\sum_{R}exp(s(r; \delta))}.
\end{aligned}
\end{equation}
Pseudo label $\hat y$ for each selection branch is selected from the  top-scoring proposals in previous stage.
% \begin{equation}
% \hat y_r =\left\{     
%              \begin{array}{ll}
%              1, & \text{if} \ p{(r;\delta)}>\tau  \\
%              0, & \text{otherwise}   
%              \end{array}.
% \right.
% \end{equation}
% The positive instance $r^\ast$ is selected by $\hat y_r = 1$. 

\textbf{Detection Branch.}Since we get pseudo labels, each proposal now has a bounding-box regression target  and  classification target. As a consequence, Detection Loss can be defined as:
\begin{equation}
\begin{aligned}
    \mathcal{L}_{det}= \mathcal{L}_{refine}+\lambda \mathcal{L}_{regression} \ ,\\
\end{aligned}
\end{equation}
where $\mathcal{L}_{refine}$ is the refine classification loss; and $\mathcal{L}_{regression}$ is bounding box regression loss. $\lambda$ is used as a weight to balance the two losses.  During the learning, a object detector is learned to generate instance bags by using the refine loss defined as:
\begin{equation}
\begin{aligned}
    \mathcal{L}_{refine}=  -\sum_{r^\ast} \log p(r^\ast,\delta) \ ,\\
\end{aligned}
\end{equation}
where $p(r^\ast,\delta)$ prediction score of the pseudo object with soft-max operation. For bounding box regression loss, smooth-L1 loss is adopted:
\begin{equation}
\begin{aligned}
    \mathcal{L}_{regression}=  \dfrac{1}{N}\sum_{r} \mathcal{L}_{smooth_{L1}}(Target(r),Box(r)) \ ,\\
\end{aligned}
\end{equation}
where $Box(r)$ is the predicted box for proposal r, and $Target(r)$ is the regression target generated by pseudo label.

\subsection{DS-MIL method}

% By revisiting the MIL-based WSOD, we derive that feature learning and proposal selection are the major factors influence the performance of network. 
% Moreover, previous methods only consider optimize the both factors together, ignoring the difference between them. As a consequence, we decide to add a proposal-wise attention branch which is parallel to the object discovery and localization branch to select key proposals. At the same time, we introduce a channel-wise Self-Attention Module on top of the base network to re-weight features CNNs learned. 

Optimizing the non-convex loss function and performing instance selection still remain to be elaborated in WSOD approaches. In C-MIL~\cite{wan2019c}, a continuation strategy is used in MIL to alleviate these two problems. 
% In C-MIL\cite{wan2019c}, it concludes that two problems, the method to optimize the non-convex loss function and the approach to perform instance selection, remain to be elaborated. And they propose a continuation method to improve it. 
However, C-MIL is still easy to be stuck into local minima because the parameters are hard to choose and the optimization strategy is complex. As a consequence, we decide to propose a novel training strategy to solve these problems. We recognize WSOD as a Discovering-and-Selection process, and design the Discovering Module and Selection Module to model this process, as shown in Fig. \ref{fig:2}.

\subsubsection{Discovey}

Dealing with localization ambiguity under only classical convolution layers is difficult, in which   the high responses are focus on the most discriminative part, therefore only a few instances are mined.  As a consequence, we propose to integrate a Discovery module into the network to capture more context information and enforce the feature learning to learn complete object feature. That means this module could help us discover more important instances.  Following \cite{wang2018non}, $u$ and $v$ denote input and output feature, $m$ and $n$ are corresponding spatial position index, the general self-attention mechanism is defined as: 
\begin{equation}
    v_m =\dfrac{1}{C(u_m)}\sum_{\forall n}f(u_m,u_n)g(u_n)+u_m, 
\end{equation}
 The output signal is normalized by $C(u_m) =  \sum_{\forall n}f(u_m,u_n)$. Function $g(u_n)$ gives a representation of input signal $u_n$ at each position and all of them are aggregated into position $m$ with the similarity weights given by $f(u_m,u_n)={\varphi(u_m)^T\varphi(u_n)}$, which calculates the dot-product pixel affinity in an embedding space.
Here, we take the inner-product to calculate the affinity between channels and integrate the similarity weights into Eq.13. %\cite{gottlob:nonmon}%
\begin{equation}
    v_m =\dfrac{1}{C(u_m)}\sum_{\forall n}Softmax({\varphi(u_m)^T\varphi(u_n)})\hat v_n,
\end{equation}
where $\hat v_n$ is the original feature map. And the similarities are activated by Softmax. The final feature map is the weighted sum of the original feature map with normalized similarities. For the final feature map, because each pixel of it is generated by a weighted sum of all pixels, more areas will be activated, part-domination could be largely improved. The self-attention module structure is illustrated in Fig. \ref{fig:2}. Compared to other self-attention methods, our proposed self-attention method has two differences: Firstly, we implement self-attention method on instance-level, which can avoid instance level feature map mixing other information and save a lot of computation capacity. Secondly, we cancel the residual connection to avoid changing the activation intensity.

% In the end, as Fig. \ref{fig:2} depicts, we plug the self-attention module before the classification branch of the network, and these two parts form our discovery module. This module can discover rich local minima and provide them for our selection module to choose.  
\begin{figure*}[!ht]
\centering
\includegraphics[width=0.99\linewidth]{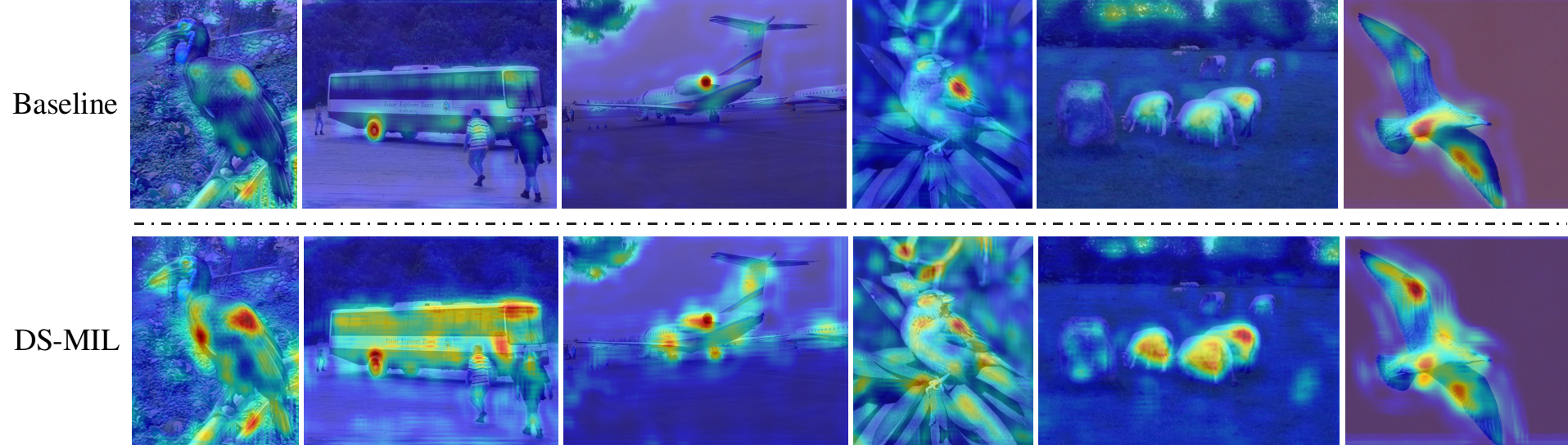}
\caption{Heatmaps of baseline and our method depict the effectiveness of Discovery Module. The heatmaps of baselines shows that baseline method only activate discriminative region for classification. The heatmaps of DS-MIL verify that DS-MIL could activate the full object region.  }
\label{fig:4}
\end{figure*}

\subsubsection{Selection}

Inaccurate classification score for proposals easily cause the localization ambiguity, e.g., Proposals cover only part of object have higher score. 
We propose a selection branch to find the confident proposal from the proposal pool produced by discovery module, which is inspired by \cite{luo2020weakly}. 

From the MIL setting, the proposals cover object determine the label of an image, while the proposals only cover background can not affect the label of an image. The proposal is regarded as key proposals when it covers the object in the image. A binary variable $h_{i}\in \{0,1\}$ is used to indicate whether proposal $R_i$ is significant for the generation of the image's label  and one predictor $\theta$ to predict the probability of a proposal belonging to different categories. Then the image's category probability can be generated:
\begin{equation}
\label{eq:p}
    \begin{aligned}
        % {\theta^*, h^*} = & \mathop{\arg\max}_{\theta, h} p_\theta(y_c=1|R,h) \\ 
        % & =\mathop{\arg\max}_{\theta, h} p_\theta(y_{c,i}=1|R_i)\cdot[h_i=1]\\
        p_\theta(y_c=1|R,h) = \mathop{\max} \{p_\theta(y_{c,i}=1|R_i)\cdot[h_i=1] \}\
    \end{aligned}
    \quad ,
\end{equation}
where $p_\theta(y_{c,i}=1|R_i)$ represents the probability that proposal $R_i$ is classified to the $c$-th category. And $p_\theta(y_c=1|R,h)$ is the probability that the image belongs to $c$-th category.
After that, We use one estimator $\phi$ to estimate $h$:
\begin{equation}
\label{eq:h}
\begin{small}
\begin{aligned}
     h = q_\phi(h|R)\\
\end{aligned}
\quad ,
\end{small}
\end{equation}

% As $h$ is a latent variable, and the image label is determined by the most important proposal's label
% \begin{equation}
% \begin{small}
% \begin{aligned}
%      p_c = & \mathop{\max} q_\phi(h|R)\cdot p_\theta(h|R,y_c)\\
% \end{aligned}
% \quad ,
% \end{small}
% \end{equation}

Then we can combine Eq.\ref{eq:p} and Eq.\ref{eq:h} and form a loss function to select the best sample among the local minima using cross entropy:
\begin{equation}
\begin{aligned}
    \mathcal{L}_{sel} = - \sum_{c=1}^C\{y_c \log(p_\theta)+(1-y_c)\log(1-p_\theta)\},\\
\end{aligned}
\end{equation}
where $p_\theta$ is $p_\theta(y_c=1|R,h)$

% In the end, as Fig.\ref{fig:2} shows, a selection module which is composed by detection branches and a selection branch is proposed. We fuse the results of the selection branch(yellow cuboid in Fig. \ref{fig:2}) and detection branch by class-wise addition. And the selection module can re-rank the confidence of proposals and select the best sample in rich local minima.

\subsubsection{Overall}
The discovery module finds multiple local minima and generate accurate pseudo labels, which is plug-and-play and no new loss is added by it. The selection module re-ranks proposals with confidence and selects the best sample with a new loss $\mathcal{L}_{sel}$. Thus, the final loss of DS-MIL is defined as:
\begin{equation}
    \begin{aligned}
        \mathcal{L} = \mathcal{L}_{cls}+\mathcal{L}_{det}+\nu\mathcal{L}_{sel}
    \end{aligned}
\end{equation}
where $\nu$ is a weight to balance the three losses.

\section{Experiment}
\subsection{Datasets and Evaluation Metrics}

\begin{table}[!ht]
\setlength{\tabcolsep}{5mm}
\centering
\caption{Ablation study of DS-MIL with Discovering Module (D) and Selection Module (S).}
\begin{tabular}{lrr}
\toprule[1 pt]
Method      & Dataset &$mAP$ \\
\midrule
MIST ~\cite{ren2020instance}       & VOC2007      & 51.4        \\
\midrule
MIST +S      & VOC2007      & 53.5         \\
MIST+D    & VOC2007      & 53.8         \\
MIST+S+D  & VOC2007      & \textbf{55.5}         \\
\bottomrule[1 pt]
\end{tabular}
\label{table:1}
\end{table}

\begin{table}[!ht]
\setlength{\tabcolsep}{7mm}
\centering
\caption{Ablation study of DS-MIL with different baselines. }
\begin{tabular}{lrr}
\toprule
Method         & Dataset     & $mAP$           \\
\midrule
OICR~\cite{tang2017multiple}          & VOC2007     & 41.2            \\
% OICR+PA        & VOC2007     & 43.3            \\
% OICR+CA        & VOC2007     & 42.0            \\
OICR+S+D     & VOC2007     & \textbf{46.5}            \\
\midrule[1 pt]
PCL~\cite{tang2018pcl}            & VOC2007     & 43.5            \\
% PCL+PA         & VOC2007     & 45.1            \\
% PCL+CA         & VOC2007     & 44.3            \\
PCL+S+D      & VOC2007     & \textbf{47.1}            \\
\bottomrule
\end{tabular}
\label{table:2}
\end{table}

\begin{table}[!ht]
\setlength{\tabcolsep}{7mm}
\centering
\caption{Ablation study of the Number of detection branches (K).}
\begin{tabular}{lrr}
\toprule[1 pt]
Method      & K &$mAP$ \\
\midrule
DS-MIL    & 1     & 51.1         \\
DS-MIL  & 2       & 54.2         \\
DS-MIL   & 3      & \textbf{55.5}         \\
DS-MIL  & 4       & 53.5         \\
\bottomrule[1 pt]
\end{tabular}
\label{table:3}
\end{table}
In experiment, we evaluate our approach on three popular datasets: PASCAL VOC 2007\&2012\cite{everingham2010pascal} and MS-COCO\cite{lin2014microsoft}. PASCAL VOC 2007\&2012 datasets\cite{everingham2010pascal} have 9962 and 22531 images of 20 object classes respectively. Only image-level annotations are used as supervision in all experiments. For PASCALC VOC, we use the trainval set(5011 images for 2007 and 11540 for 2012) for training and test set for testing. For evaluation on PASCAL VOC, two metrics are used to evaluate our model. First, we evaluate detection performance using mean Average Precision (mAP) on the PASCAL VOC 2007 and 2012 test set. Second, we evaluate the localization accuracy using Correct Localization (CorLoc) on PASCAL VOC 2007 and 2012 trainval set. Based on the PASCAL criterion, a predicted box is considered positive if it has an IoU $>$ 0.5 with a ground-truth bounding box. MS-COCO\cite{lin2014microsoft} contains 80 categories. We train on train2017 split and evaluate on val2017 split, which consists of 118287 and 5000 images, respectively. $mAP_{0.5}$ (IoU threshold at 0.5) and $mAP$ (averaged over IoU thresholds in [0.5 : 0.05 : 0.95]) on val2017 are reported.
\subsection{Implementation Details} 

VGG16 \cite{simonyan2014very} pre-trained on ImageNet \cite{russakovsky2015imagenet} is used as the backbone in experiment. Selective Search\cite{uijlings2013selective} is used to generate about 2,000 proposals per-image for PASCAL VOC and MCG is used for MS-COCO. The maximum iteration numbers are set to be 150k, 160k and 300k for VOC 2007, VOC 2012 and MS-COCO respectively. The whole WSOD network is by stochastic gradient descent (SGD) with a momentum of 0.9, an initial learning rate of 0.001 and a weight decay of 0.0005. The learning rate will decay with a factor of 10 at the 75kth, 80kth and 150kth iterations for VOC 2007, VOC 2012 and MS-COCO, respectively. The total number of refinement branches is set to be 3. For data augmentation, we use six image scales $\{480, 576, 688, 864, 1000, 1200\}$ (resize the shortest side to one of these scales) and cap the longest image side to less than 2000 with horizontal flips for both training and testing. The balance weights $\lambda$ and $\nu$ is 0.05 and 0.1.
\begin{table*}[!t]
\centering
\caption{The object detection performance comparison with state-of-the-art on PASCAL VOC 2007 TEST SET (VGG16 backbone, $mAP_{0.5}$). }
% \footnotesize
\resizebox{\textwidth}{28mm}
{
\begin{tabular}{lccccccccccccccccccccc}
\toprule
Method   & areo & bike & bird & boat & bottle & bus & car & cat & chair & cow & table & dog & horse & motor & person & plant & sheep & sofa & train & tv  & $mAP$  \\
\midrule
WSDDN\cite{Bilen_2016_CVPR}   &46.4 &58.1 &35.5 &25.9 &14.0 &66.7 &53.0 &39.2 &8.9 &41.8 &26.6 &38.6 &44.7 &59.0 &10.8 &17.3 &40.7 &49.6 &56.9 &50.8    & 34.8         \\
OICR\cite{tang2017multiple}    &58.5 &63.0 &35.1 &16.9 &17.4 &63.2 &60.8 &34.4 &8.2 &49.7 &41.0 &31.3 &51.9 &64.8 &13.6 &23.1 &41.6 &48.4 &58.9 &58.7       & 41.2   \\
DSTL\cite{jie2017deep}    &52.2 &47.1 &35.0 &26.7 &15.4 &61.3 &66.0 &54.3 &3.0 &53.6 &24.7 &43.6 &48.4 &65.8 &6.6 &18.8 &51.9 &43.6 &53.6 &62.4       &  41.7   \\
WCCN\cite{diba2017weakly}    &49.5 &60.6 &38.6 &29.2 &16.2 &70.8 &56.9 &42.5 &10.9 &44.1 &29.9 &42.2 &47.9 &64.1 &13.8 &23.5 &45.9 &54.1 &60.8 &54.5       & 42.8   \\
PCL\cite{tang2018pcl}     &57.1 &67.1 &40.9 &16.9 &18.8 &65.1 &63.7 &45.3 &17.0 &56.7 &48.9 &33.2 &54.4 &68.3 &16.8 &25.7 &45.8 &52.2 &59.1 &62.0       & 43.5   \\
ZLDN\cite{zhang2018zigzag}    &55.4 &68.5 &50.1 &16.8 &20.8 &62.7 &66.8 &56.5 &2.1 &57.8 &47.5 &40.1 &69.7 &68.2 &21.6 &27.2 &53.4 &56.1 &52.5 &58.2       & 47.6   \\
MELM\cite{wan2018min}    &55.6 &66.9 &34.2 &29.1 &16.4 &68.8 &68.1 &43.0 &25.0 &65.6 &45.3 &53.2 &49.6 &68.6 &2.0 &25.4 &52.5 &56.8 &62.1 &57.1        & 47.3   \\
C-WSL\cite{gao2018c}   &62.7 &63.7 &40.0 &25.5 &17.7 &70.1 &68.3 &38.9 &25.4 &54.5 &41.6 &29.9 &37.9 &64.2 &11.3 &27.4 &49.3 &54.7 &61.4 &67.4       & 45.6   \\
WSPRN\cite{tang2018weakly}   &57.9 &70.5 &37.8 &5.7 &21.0 &66.1 &69.2 &59.4 &3.4 &57.1 &57.3 &35.2 &64.2 &68.6  &32.8 &28.6 &50.8 &49.5 &41.1 &30.0       & 45.3   \\
C-MIL\cite{wan2019c}   &62.5 &58.4 &49.5 &32.1 &19.8 &70.5 &66.1 &63.4 &20.0 &60.5 &52.9 &53.5 &57.4 &68.9 &8.4 &24.6 &51.8 &58.7 &66.7 &63.6        & 50.5   \\
WSOD2\cite{zeng2019wsod2}   &65.1 &64.8 &57.2 &39.2 &24.3 &69.8 &66.2 &61.0 &29.8 &64.6 &42.5 &60.1 &71.2 &70.7 &21.9 &28.1 &58.6 &59.7 &52.2 &64.8       & 53.6   \\
C-MIDN\cite{gao2019c}  &53.3 &71.5 &49.8 &26.1 &20.3 &70.3 &69.9 &68.3 &28.7 &65.3 &45.1 &64.6 &58.0 &71.2 &20.0 &27.5 &54.9 &54.9 &69.4 &63.5       &52.6   \\
\midrule
DS-MIL(ours)       &\textbf{66.9} &\textbf{78.2} &52.5 &29.0 &\textbf{27.0} &68.6 &\textbf{77.0} &\textbf{71.9} &29.5 &\textbf{71.9} &46.8 &61.9 &58.3 &\textbf{75.6} &\textbf{32.8} &26.7 &\textbf{62.9} &53.1 &50.4 &\textbf{69.6}         &\textbf{55.5}          \\
\bottomrule
\end{tabular}
}
\label{table:4}
\end{table*}
\begin{table*}
\centering
\caption{The Object Localization performance comparison with state-of-the-art on the PASCAL VOC 2007 TRAINVAL SET (VGG16 backbone, $Corloc$). }
% \footnotesize
\resizebox{\textwidth}{28mm}
{
\begin{tabular}{lccccccccccccccccccccc}
\toprule
Method     & areo & bike & bird & boat & bottle & bus & car & cat & chair & cow & table & dog & horse & motor & person & plant & sheep & sofa & train & tv & $mAP$  \\
\midrule
WSDDN\cite{Bilen_2016_CVPR}   &65.1 &58.8 &58.5 &33.1 &39.8 &68.3 &60.2 &59.6 &34.8 &64.5 &30.5 &43.0 &56.8 &82.4 &25.5 &41.6 &61.5 &55.9 &65.9 &63.7    & 53.5         \\
OICR\cite{tang2017multiple}    &81.7 &80.4 &48.7 &49.5 &32.8 &81.7 &85.4 &40.1 &40.6 &79.5 &35.7 &33.7 &60.5 &88.8 &21.8 &57.9 &76.3 &59.9 &75.3 &81.4       & 60.6   \\
DSTL\cite{jie2017deep}    &72.7 &55.3 &53.0 &27.8 &35.2 &68.6 &81.9 &60.7 &11.6 &71.6 &29.7 &54.3 &64.3 &88.2 &22.2 &53.7 &72.2 &52.6 &68.9 &75.5       & 56.1   \\
WCCN\cite{diba2017weakly}    &83.9 &72.8 &64.5 &44.1 &40.1 &65.7 &82.5 &58.9 &33.7 &72.5 &25.6 &53.7 &67.4 &77.4 &26.8 &49.1 &68.1 &27.9 &64.5 &55.7       & 56.7   \\
PCL\cite{tang2018pcl}     &79.6 &85.5 &62.2 &47.9 &37.0 &83.8 &83.4 &43.0 &38.3 &80.1 &50.6 &30.9 &57.8 &90.8 &27.0 &58.2 &75.3 &68.5 &75.7 &78.9       & 62.7   \\
ZLDN\cite{zhang2018zigzag}     &74.0 &77.8 &65.2 &37.0 &46.7 &75.8 &83.7 &58.8 &17.5 &73.1 &49.0 &51.3 &76.7 &87.4 &30.6 &47.8 &75.0 &62.5 &64.8 &68.8       & 61.2   \\
MELM\cite{wan2018min}     &- &- &- &- &- &- &- &- &- &- &- &- &- &- &- &- &- &- &- &-       & 61.4  \\
C-WSL\cite{gao2018c}    &86.3 &80.4 &58.3 &50.0 &36.6 &85.8 &86.2 &47.1 &42.7 &81.5 &42.2 &42.6 &50.7 &90.0 &14.3 &61.9 &85.6 &64.2 &77.2 &82.4       & 63.3  \\
WSRPN\cite{tang2018weakly}    &77.5 &81.2 &55.3 &19.7 &44.3 &80.2 &86.6 &69.5 &10.1 &87.7 &68.4 &52.1 &84.4 &91.6 &57.4 &63.4 &77.3 &58.1 &57.0 &53.8       & 63.8  \\
C-MIL\cite{wan2019c}     &- &- &- &- &- &- &- &- &- &- &- &- &- &- &- &- &- &- &- &-       & 65.0  \\
WSOD2\cite{zeng2019wsod2}   &87.1 &80.0 &74.8 &60.1 &36.6 &79.2 &83.8 &70.6 &43.5 &88.4 &46.0 &74.7 &87.4 &90.8 &44.2 &52.4 &81.4 &61.8 &67.7 &79.9       & 69.5   \\
C-MIDN\cite{gao2019c}  &- &- &- &- &- &- &- &- &- &- &- &- &- &- &- &- &- &- &- &-       &68.7   \\
\midrule
DS-MIL(ours)       &85.3 &84.5 &\textbf{78.1} &51.5 &\textbf{48.4} &\textbf{83.6} &\textbf{88.1} &\textbf{72.7} &\textbf{50.1} &80.8 &48.5 &52.1 &72.0 &\textbf{92.4} &43.7 &55.2 &80.5 &54.3 &\textbf{77.8} &79.7         &69.0          \\
\bottomrule
\end{tabular}
}
\label{table:5}
\end{table*}

\subsection{Ablation Study}
We conduct ablation experiments on PASCAL VOC 2007 to prove the effectiveness of our proposed DS-MIL approach from 4 perspectives.

% \begin{figure*}[t]
% \centering
% \includegraphics[scale=1.2]{failure.png}
% \caption{Failure cases of DS-MIL approach.In the first column, multiple objects are recognised as single object. In the second column, absence of boxes means no object is detected. In the third column, objects is missing. }
% \label{fig:6}
% \end{figure*}

% \textbf{Discovering Module.} 
For discovery module, we adopt MIST without Concrete DropBlock(CDB)~\cite{ren2020instance} as our baseline to verify its effectiveness. We add a single discovering module on the baseline, as shown in Table \ref{table:1}, the performance is improved to 53.8\%, which indicates one discovering module is effective.
% \textbf{Selection Module.} 
To verify the effect of our newly proposed selection module, we use the same baseline with discovery module.  As Table \ref{table:1} depicted, the selection module improves the performance of baseline by 2.1\%.
As bounding box regressor is plugged into several WSOD approach and illustrated that it's effective for performance gain \cite{gao2018c}. Following \cite{gao2018c}, we also add the regressor to the proposed approach, and we achieves 55.5\% on PASCAL VOC 2007 by adding a discovery module and a selection module.

% \textbf{Discovering Module and Selection Module}
As the discovery module and selection module are plug and play, we conduct experiments with other two baselines, $i.e.$, OICR~\cite{tang2017multiple} and PCL~\cite{tang2018pcl}.
The results verify that our method gain improvements in all of three baselines. For each baseline, a selection module and a discovery module are added. In Table \ref{table:2}, the $mAP$ performance increases 5.3\% for OICR and 3.6\% for PCL.

\begin{table}
\setlength{\tabcolsep}{10mm}
\centering
\caption{The Object detection performance on the PASCAL VOC 2012 TEST SET (VGG16 backbone, $mAP_{0.5}$). }
% \footnotesize
{
\begin{tabular}{lrr}
\toprule
Method   & $mAP$  \\
\midrule
OICR\cite{tang2017multiple}          & 37.9   \\
PCL\cite{tang2018pcl}           & 40.6   \\
C-MIL\cite{wan2019c}         & 46.7   \\
WSOD2\cite{zeng2019wsod2}         & 47.2   \\
C-MIDN\cite{gao2019c}        &50.2   \\
\midrule
DS-MIL(ours)    &\textbf{50.4}          \\
\bottomrule
\end{tabular}
}
\label{table:6}
\end{table}
\begin{figure*}[t]
\centering
\includegraphics[scale=0.5]{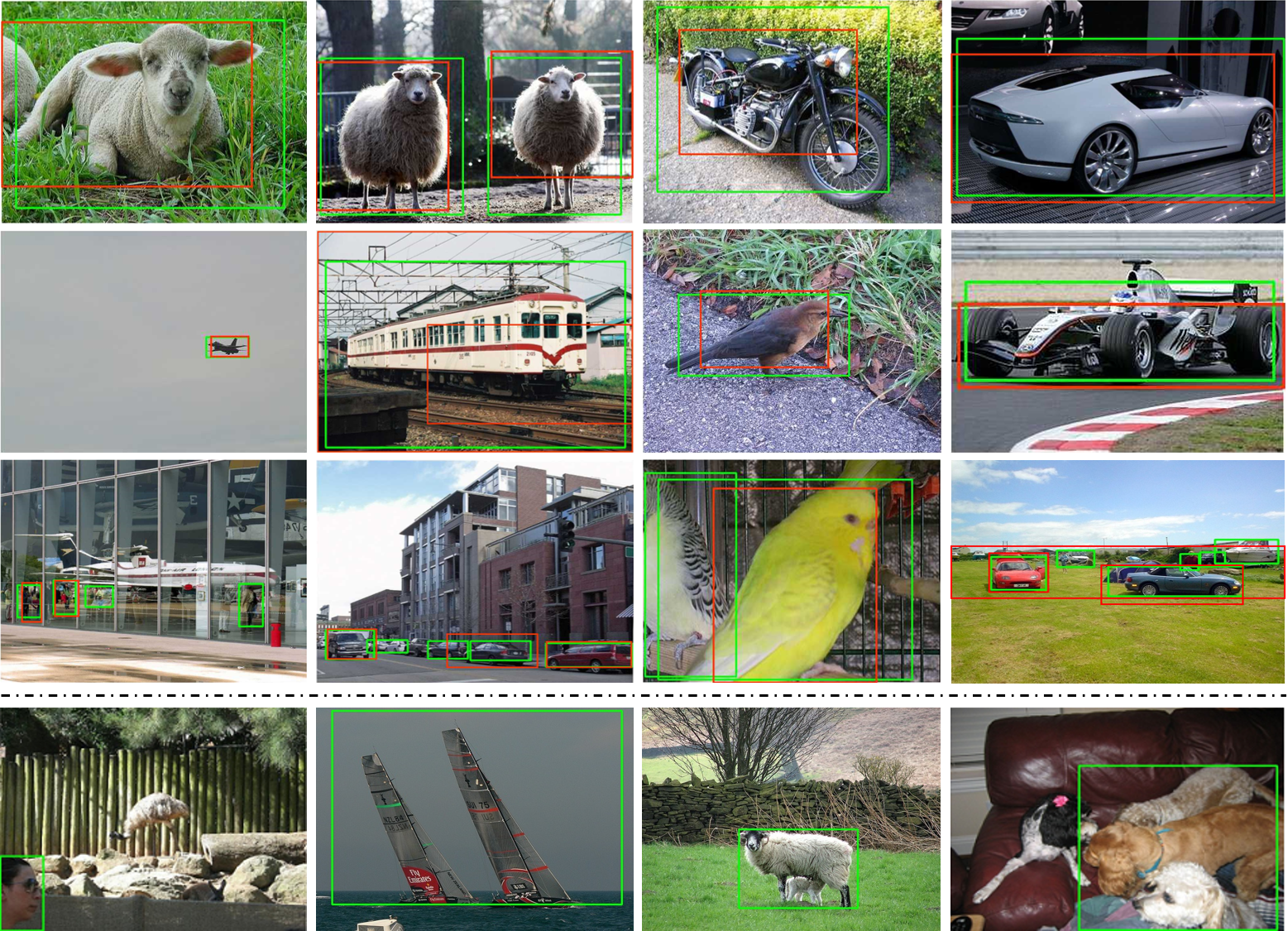}
\caption{Visualization of DS-MIL results and the baseline. In upper part, the results of DS-MIL are shown in green boxes and the results of the baseline are shown in red boxes. In bottom part, some of failure cases of DS-MIL are shown. }
\label{fig:5}
\end{figure*}
\begin{table}
\setlength{\tabcolsep}{10mm}
\centering
\caption{The Object Localization performance on the PASCAL VOC 2012 TRAINVAL SET (VGG16 backbone, $Corloc$) . }
% \footnotesize
{
\begin{tabular}{lrr}
\toprule
Method     & $mAP$  \\
\midrule
OICR\cite{tang2017multiple}          & 62.1   \\
PCL\cite{tang2018pcl}            & 63.2   \\
C-MIL\cite{wan2019c}          & 67.4   \\
WSOD2\cite{zeng2019wsod2}          & \textbf{71.9}   \\
C-MIDN\cite{gao2019c}        & 71.2   \\
\midrule
DS-MIL(ours)              &70.6         \\
\bottomrule
\end{tabular}
}
\label{table:7}
\end{table}

% \textbf{Visualization}
In Fig. \ref{fig:4}, we provide some comparisons between the heatmaps of baseline and our approach. Obviously, the baseline activated the discriminated regions but ignore full object extent. Compared to the baseline, DS-MIL shows great performance by activating more regions to cover the whole object. The main reason accounts for this result is our discovery module could capture more object extent and provide more accurate object localization information for detectors. On the contrary, baseline method only considers object classification and hardly optimizes object localization.

The number of detection branches determines how many times we refine the detection results. We also conduct some experiments on it.  The number of branches is set to be K, and four different Ks: 1,2,3,4 are adopted. While we change the value of K, the rest of the hyper-parameters are fixed.  Table \ref{table:3} shows the influence of K. We can find that when K is set to be 1, the mAP is only 51.1\%. Then, the performance becomes better with the increasing of K. When K is set to be 3, it achieves the best performance which is 55.5\%. And the result decreases when the K is equal to 4. The reason is those chosen proposals are too scattered for the 4th branch.

% \textbf{Time cost}
% Our training time is nearly $2\times$ longer than the baseline MELM\cite{wan2018min}, and the testing speed of our method is nearly $2\times$ slower than the baseline.

\subsection{Comparison with Other Methods}

\textbf{VOC dataset}
In this comparison, we adopt MIST as our baseline, and add one discovery module and one selection module to the baseline network. Besides, we use bounding box regressor to revise the location of the predicted boxes. In order to verify the effectiveness of DS-MIL, 12 state-of-the-art WSOD methods are compared with our method and most of the chosen methods are published in the last two years. To fully compare with these methods, we report both mAP results and Corloc results on VOC 2007 and VOC 2012 datasets are shown in Table \ref{table:4}, Table \ref{table:5}, Table \ref{table:6} and Table \ref{table:7}. From the Table \ref{table:4}, we can see that our method  outperforms all methods on VOC2007 dataset and achieves the highest mAP performance on 10 out of 20 categories. From Table\ref{table:5}, the result is little lower than state-of-the-art methods, but our method also achieves best performance on 8 classes. Table \ref{table:6} and Table \ref{table:7} shows the competitive results achieved by our method on VOC2012, it is noteworthy that our proposed method outperforms 5 previous methods.

\textbf{MS-COCO dataset}
MS-COCO is larger dataset compared to PASCAL VOC, and only few previous approaches report results on it for the difficulty of obtaining good results on it. We report our results in Table \ref{table:8}. We can find that our proposed approach achieves 12.3\% for $mAP$ and 24.5\% for $mAP_{0.5}$ which significantly outperforms previous works.

\begin{table}
\centering
\caption{Comparison with SOTA WSOD results on MS-COCO (VGG16 backbone, $Average Precision$).}
\setlength{\tabcolsep}{8mm}
{
\begin{tabular}{lrr}
\toprule
Method     & $mAP$ & $mAP_{0.5}$  \\
\midrule
WSDDN\cite{Bilen_2016_CVPR}   &-     &11.5   \\
MELM\cite{wan2018min}         &-     &18.8   \\
PCL\cite{tang2018pcl}         &8.5   &19.4   \\
PG-PS\cite{cheng2020high}     &-     &20.7   \\
C-MIDN\cite{gao2019c}         &9.4   &21.4   \\
WSOD2\cite{zeng2019wsod2}     &10.8  &22.7   \\
\midrule
DS-MIL(ours)                    &\textbf{12.3}      &\textbf{24.5}         \\
\bottomrule
\end{tabular}
}
\label{table:8}
\end{table}

\subsection{Visualization}

In Fig. \ref{fig:5}, we visualize some detection results of our proposed method and the baseline approach. The green boxes represent DS-MIL results and the red boxes represent the baseline, respectively. The first two rows of Fig. \ref{fig:5} proves that our proposed approach largely improves the part-dominant problem and the third row of Fig. \ref{fig:5} shows DS-MIL has the better capability to detect multiple objects. As a consequence, we can conclude that DS-MIL performs much better than the baseline. Moreover, the visualization results also shows that our approach tends to cover more extent of objects and avoid selecting incomplete proposals. And these are the effects of Selection Module and Discovering Module. In the last row of Fig. \ref{fig:5}, we also show some failure cases of our method. As we can see, our detector will recognize multiple objects as single object or miss some objects. These failures are come from two factors: (1) The occlusion of objects. (2)The Selective Search algorithm\cite{uijlings2013selective} may not generate good proposal. And we believe these problems could be improved by applying network with stronger representation ability (e.g. transformer based network) or combining with Class Activation Map.

\section{Conclusion}
In this paper, We proposed an effective and novel method, referred to as DS-MIL, for weakly supervised object detection. DS-MIL targets alleviating the part-dominant problem of multiple instance learning using a new training strategy: discovering-and-selection. This strategy is achieved by introducing a Discovering Module and a Selection Module. DS-MIL significantly improved performance of weakly supervised object detection on PASCAL VOC 2007, PASCAL VOC 2012 and MS-COCO datasets.
\section*{Acknowledgements}\label{}
This work is funded by National Key Research and Development Project of China under Grant No. 2020AAA0105600, National Natural Science Foundation of China under Grant No. 62006182, and the Fundamental Research Funds for the Central Universities under Grant No. xzy012020017.


\bibliographystyle{elsarticle-num}
\bibliography{refs}

\begin{thebibliography}{10}
\expandafter\ifx\csname url\endcsname\relax
  \def\url#1{\texttt{#1}}\fi
\expandafter\ifx\csname urlprefix\endcsname\relax\def\urlprefix{URL }\fi
\expandafter\ifx\csname href\endcsname\relax
  \def\href#1#2{#2} \def\path#1{#1}\fi

\bibitem{Bilen_2016_CVPR}
H.~Bilen, A.~Vedaldi, Weakly supervised deep detection networks, in:
  Proceedings of the IEEE Conference on Computer Vision and Pattern Recognition
  (CVPR), 2016.

\bibitem{diba2017weakly}
A.~Diba, V.~Sharma, A.~Pazandeh, H.~Pirsiavash, L.~Van~Gool, Weakly supervised
  cascaded convolutional networks, in: Proceedings of the IEEE conference on
  computer vision and pattern recognition, 2017, pp. 914--922.

\bibitem{tang2017multiple}
P.~Tang, X.~Wang, X.~Bai, W.~Liu, Multiple instance detection network with
  online instance classifier refinement, in: Proceedings of the IEEE Conference
  on Computer Vision and Pattern Recognition, 2017, pp. 2843--2851.

\bibitem{tang2018pcl}
P.~Tang, X.~Wang, S.~Bai, W.~Shen, X.~Bai, W.~Liu, A.~Yuille, Pcl: Proposal
  cluster learning for weakly supervised object detection, IEEE transactions on
  pattern analysis and machine intelligence 42~(1) (2018) 176--191.

\bibitem{wan2018min}
F.~Wan, P.~Wei, J.~Jiao, Z.~Han, Q.~Ye, Min-entropy latent model for weakly
  supervised object detection, in: Proceedings of the IEEE Conference on
  Computer Vision and Pattern Recognition, 2018, pp. 1297--1306.

\bibitem{wan2019c}
F.~Wan, C.~Liu, W.~Ke, X.~Ji, J.~Jiao, Q.~Ye, C-mil: Continuation multiple
  instance learning for weakly supervised object detection, in: Proceedings of
  the IEEE Conference on Computer Vision and Pattern Recognition, 2019, pp.
  2199--2208.

\bibitem{huang2020comprehensive}
Z.~Huang, Y.~Zou, V.~Bhagavatula, D.~Huang, Comprehensive attention
  self-distillation for weakly-supervised object detection, arXiv preprint
  arXiv:2010.12023 (2020).

\bibitem{DBLP:conf/nips/MaronL97}
O.~Maron, T.~Lozano-Pérez, A framework for multiple-instance learning, in:
  NIPS, 1997, pp. 570--576.

\bibitem{DBLP:conf/cvpr/HoffmanPDS15}
J.~Hoffman, D.~Pathak, T.~Darrell, K.~Saenko, Detector discovery in the wild:
  Joint multiple instance and representation learning, in: {IEEE} Conference on
  Computer Vision and Pattern Recognition, {CVPR} 2015, Boston, MA, USA, June
  7-12, 2015, 2015, pp. 2883--2891.

\bibitem{DBLP:journals/pami/CinbisVS17}
R.~G. Cinbis, J.~J. Verbeek, C.~Schmid, Weakly supervised object localization
  with multi-fold multiple instance learning, {IEEE} Trans. Pattern Anal. Mach.
  Intell. 39~(1) (2017) 189--203.

\bibitem{ren2020instance}
Z.~Ren, Z.~Yu, X.~Yang, M.-Y. Liu, Y.~J. Lee, A.~G. Schwing, J.~Kautz,
  Instance-aware, context-focused, and memory-efficient weakly supervised
  object detection, in: Proceedings of the IEEE/CVF Conference on Computer
  Vision and Pattern Recognition, 2020, pp. 10598--10607.

\bibitem{li2016weakly}
D.~Li, J.-B. Huang, Y.~Li, S.~Wang, M.-H. Yang, Weakly supervised object
  localization with progressive domain adaptation, in: Proceedings of the IEEE
  Conference on Computer Vision and Pattern Recognition, 2016, pp. 3512--3520.

\bibitem{wang2018non}
X.~Wang, R.~Girshick, A.~Gupta, K.~He, Non-local neural networks, in:
  Proceedings of the IEEE conference on computer vision and pattern
  recognition, 2018, pp. 7794--7803.

\bibitem{uijlings2013selective}
J.~R. Uijlings, K.~E. Van De~Sande, T.~Gevers, A.~W. Smeulders, Selective
  search for object recognition, International journal of computer vision
  104~(2) (2013) 154--171.

\bibitem{singh2016track}
K.~K. Singh, F.~Xiao, Y.~J. Lee, Track and transfer: Watching videos to
  simulate strong human supervision for weakly-supervised object detection, in:
  Proceedings of the IEEE Conference on Computer Vision and Pattern
  Recognition, 2016, pp. 3548--3556.

\bibitem{ge2018multi}
W.~Ge, S.~Yang, Y.~Yu, Multi-evidence filtering and fusion for multi-label
  classification, object detection and semantic segmentation based on weakly
  supervised learning, in: Proceedings of the IEEE Conference on Computer
  Vision and Pattern Recognition, 2018, pp. 1277--1286.

\bibitem{kantorov2016contextlocnet}
V.~Kantorov, M.~Oquab, M.~Cho, I.~Laptev, Contextlocnet: Context-aware deep
  network models for weakly supervised localization, in: European Conference on
  Computer Vision, Springer, 2016, pp. 350--365.

\bibitem{choe2019attention}
J.~Choe, H.~Shim, Attention-based dropout layer for weakly supervised object
  localization, in: Proceedings of the IEEE Conference on Computer Vision and
  Pattern Recognition, 2019, pp. 2219--2228.

\bibitem{shen2019cyclic}
Y.~Shen, R.~Ji, Y.~Wang, Y.~Wu, L.~Cao, Cyclic guidance for weakly supervised
  joint detection and segmentation, in: Proceedings of the IEEE Conference on
  Computer Vision and Pattern Recognition, 2019, pp. 697--707.

\bibitem{tang2018weakly}
P.~Tang, X.~Wang, A.~Wang, Y.~Yan, W.~Liu, J.~Huang, A.~Yuille, Weakly
  supervised region proposal network and object detection, in: Proceedings of
  the European conference on computer vision (ECCV), 2018, pp. 352--368.

\bibitem{wei2018ts2c}
Y.~Wei, Z.~Shen, B.~Cheng, H.~Shi, J.~Xiong, J.~Feng, T.~Huang, Ts2c: Tight box
  mining with surrounding segmentation context for weakly supervised object
  detection, in: Proceedings of the European Conference on Computer Vision
  (ECCV), 2018, pp. 434--450.

\bibitem{zhang2018w2f}
Y.~Zhang, Y.~Bai, M.~Ding, Y.~Li, B.~Ghanem, W2f: A weakly-supervised to
  fully-supervised framework for object detection, in: Proceedings of the IEEE
  conference on computer vision and pattern recognition, 2018, pp. 928--936.

\bibitem{GAO2022108233}
W.~Gao, F.~Wan, J.~Yue, S.~Xu, Q.~Ye,
  \href{https://www.sciencedirect.com/science/article/pii/S0031320321004143}{Discrepant
  multiple instance learning for weakly supervised object detection}, Pattern
  Recognition 122 (2022) 108233.
\newblock \href {https://doi.org/https://doi.org/10.1016/j.patcog.2021.108233}
  {\path{doi:https://doi.org/10.1016/j.patcog.2021.108233}}.
\newline\urlprefix\url{https://www.sciencedirect.com/science/article/pii/S0031320321004143}

\bibitem{ZHANG201868}
Y.~Zhang, Y.~Bai, M.~Ding, Y.~Li, B.~Ghanem,
  \href{https://www.sciencedirect.com/science/article/pii/S0031320318302346}{Weakly-supervised
  object detection via mining pseudo ground truth bounding-boxes}, Pattern
  Recognition 84 (2018) 68--81.
\newblock \href {https://doi.org/https://doi.org/10.1016/j.patcog.2018.07.005}
  {\path{doi:https://doi.org/10.1016/j.patcog.2018.07.005}}.
\newline\urlprefix\url{https://www.sciencedirect.com/science/article/pii/S0031320318302346}

\bibitem{zeng2019wsod2}
Z.~Zeng, B.~Liu, J.~Fu, H.~Chao, L.~Zhang, Wsod2: Learning bottom-up and
  top-down objectness distillation for weakly-supervised object detection, in:
  Proceedings of the IEEE International Conference on Computer Vision, 2019,
  pp. 8292--8300.

\bibitem{gao2018c}
M.~Gao, A.~Li, R.~Yu, V.~I. Morariu, L.~S. Davis, C-wsl: Count-guided weakly
  supervised localization, in: Proceedings of the European Conference on
  Computer Vision (ECCV), 2018, pp. 152--168.

\bibitem{gao2019c}
Y.~Gao, B.~Liu, N.~Guo, X.~Ye, F.~Wan, H.~You, D.~Fan, C-midn: Coupled multiple
  instance detection network with segmentation guidance for weakly supervised
  object detection, in: Proceedings of the IEEE International Conference on
  Computer Vision, 2019, pp. 9834--9843.

\bibitem{liu2019utilizing}
B.~Liu, Y.~Gao, N.~Guo, X.~Ye, F.~Wan, H.~You, D.~Fan, Utilizing the
  instability in weakly supervised object detection., in: CVPR Workshops, 2019.

\bibitem{jie2017deep}
Z.~Jie, Y.~Wei, X.~Jin, J.~Feng, W.~Liu, Deep self-taught learning for weakly
  supervised object localization, in: Proceedings of the IEEE Conference on
  Computer Vision and Pattern Recognition, 2017, pp. 1377--1385.

\bibitem{li2019weakly}
X.~Li, M.~Kan, S.~Shan, X.~Chen, Weakly supervised object detection with
  segmentation collaboration, in: Proceedings of the IEEE International
  Conference on Computer Vision, 2019, pp. 9735--9744.

\bibitem{wang2017untrimmednets}
L.~Wang, Y.~Xiong, D.~Lin, L.~Van~Gool, Untrimmednets for weakly supervised
  action recognition and detection, in: Proceedings of the IEEE conference on
  Computer Vision and Pattern Recognition, 2017, pp. 4325--4334.

\bibitem{nguyen2018weakly}
P.~Nguyen, T.~Liu, G.~Prasad, B.~Han, Weakly supervised action localization by
  sparse temporal pooling network, in: Proceedings of the IEEE Conference on
  Computer Vision and Pattern Recognition, 2018, pp. 6752--6761.

\bibitem{singh2017hide}
K.~K. Singh, Y.~J. Lee, Hide-and-seek: Forcing a network to be meticulous for
  weakly-supervised object and action localization, in: 2017 IEEE international
  conference on computer vision (ICCV), IEEE, 2017, pp. 3544--3553.

\bibitem{zeng2019breaking}
R.~Zeng, C.~Gan, P.~Chen, W.~Huang, Q.~Wu, M.~Tan, Breaking winner-takes-all:
  Iterative-winners-out networks for weakly supervised temporal action
  localization, IEEE Transactions on Image Processing 28~(12) (2019)
  5797--5808.

\bibitem{Liu_2019_CVPR}
D.~Liu, T.~Jiang, Y.~Wang, Completeness modeling and context separation for
  weakly supervised temporal action localization, in: Proceedings of the
  IEEE/CVF Conference on Computer Vision and Pattern Recognition (CVPR), 2019.

\bibitem{duchenne2009automatic}
O.~Duchenne, I.~Laptev, J.~Sivic, F.~Bach, J.~Ponce, Automatic annotation of
  human actions in video, in: 2009 IEEE 12th International Conference on
  Computer Vision, IEEE, 2009, pp. 1491--1498.

\bibitem{gan2016webly}
C.~Gan, C.~Sun, L.~Duan, B.~Gong, Webly-supervised video recognition by
  mutually voting for relevant web images and web video frames, in: European
  Conference on Computer Vision, Springer, 2016, pp. 849--866.

\bibitem{ding2018weakly}
L.~Ding, C.~Xu, Weakly-supervised action segmentation with iterative soft
  boundary assignment, in: Proceedings of the IEEE Conference on Computer
  Vision and Pattern Recognition, 2018, pp. 6508--6516.

\bibitem{luo2020weakly}
Z.~Luo, D.~Guillory, B.~Shi, W.~Ke, F.~Wan, T.~Darrell, H.~Xu,
  Weakly-supervised action localization with expectation-maximization
  multi-instance learning, in: European conference on computer vision,
  Springer, 2020, pp. 729--745.

\bibitem{itti2001computational}
L.~Itti, C.~Koch, Computational modelling of visual attention, Nature reviews
  neuroscience 2~(3) (2001) 194--203.

\bibitem{larochelle2010learning}
H.~Larochelle, G.~E. Hinton, Learning to combine foveal glimpses with a
  third-order boltzmann machine, in: Advances in neural information processing
  systems, 2010, pp. 1243--1251.

\bibitem{woo2018cbam}
S.~Woo, J.~Park, J.-Y. Lee, I.~So~Kweon, Cbam: Convolutional block attention
  module, in: Proceedings of the European conference on computer vision (ECCV),
  2018, pp. 3--19.

\bibitem{fu2019dual}
J.~Fu, J.~Liu, H.~Tian, Y.~Li, Y.~Bao, Z.~Fang, H.~Lu, Dual attention network
  for scene segmentation, in: Proceedings of the IEEE Conference on Computer
  Vision and Pattern Recognition, 2019, pp. 3146--3154.

\bibitem{hu2018squeeze}
J.~Hu, L.~Shen, G.~Sun, Squeeze-and-excitation networks, in: Proceedings of the
  IEEE conference on computer vision and pattern recognition, 2018, pp.
  7132--7141.

\bibitem{huang2020improving}
Z.~Huang, W.~Ke, D.~Huang, Improving object detection with inverted attention,
  in: 2020 IEEE Winter Conference on Applications of Computer Vision (WACV),
  IEEE, 2020, pp. 1294--1302.

\bibitem{ke2020multiple}
W.~Ke, T.~Zhang, Z.~Huang, Q.~Ye, J.~Liu, D.~Huang, Multiple anchor learning
  for visual object detection, in: Proceedings of the IEEE/CVF Conference on
  Computer Vision and Pattern Recognition, 2020, pp. 10206--10215.

\bibitem{hu2018relation}
H.~Hu, J.~Gu, Z.~Zhang, J.~Dai, Y.~Wei, Relation networks for object detection,
  in: Proceedings of the IEEE Conference on Computer Vision and Pattern
  Recognition, 2018, pp. 3588--3597.

\bibitem{everingham2010pascal}
M.~Everingham, L.~Van~Gool, C.~K. Williams, J.~Winn, A.~Zisserman, The pascal
  visual object classes (voc) challenge, International journal of computer
  vision 88~(2) (2010) 303--338.

\bibitem{lin2014microsoft}
T.-Y. Lin, M.~Maire, S.~Belongie, J.~Hays, P.~Perona, D.~Ramanan,
  P.~Doll{\'a}r, C.~L. Zitnick, Microsoft coco: Common objects in context, in:
  European conference on computer vision, Springer, 2014, pp. 740--755.

\bibitem{simonyan2014very}
K.~Simonyan, A.~Zisserman, Very deep convolutional networks for large-scale
  image recognition, arXiv preprint arXiv:1409.1556 (2014).

\bibitem{russakovsky2015imagenet}
O.~Russakovsky, J.~Deng, H.~Su, J.~Krause, S.~Satheesh, S.~Ma, Z.~Huang,
  A.~Karpathy, A.~Khosla, M.~Bernstein, et~al., Imagenet large scale visual
  recognition challenge, International journal of computer vision 115~(3)
  (2015) 211--252.

\bibitem{zhang2018zigzag}
X.~Zhang, J.~Feng, H.~Xiong, Q.~Tian, Zigzag learning for weakly supervised
  object detection, in: Proceedings of the IEEE Conference on Computer Vision
  and Pattern Recognition, 2018, pp. 4262--4270.

\bibitem{cheng2020high}
G.~Cheng, J.~Yang, D.~Gao, L.~Guo, J.~Han, High-quality proposals for weakly
  supervised object detection, IEEE Transactions on Image Processing 29 (2020)
  5794--5804.

\end{thebibliography}


%\begin{thebibliography}{00}
%%% Text of bibliographic item
%
%\bibitem{}
%
%\end{thebibliography}

\end{document}